\documentclass[a4paper,twoside]{article}

\usepackage{epsfig}
\usepackage{subfigure}
\usepackage{calc}
\usepackage{amssymb}
\usepackage{amstext}
\usepackage{amsmath}
\usepackage{amsthm}
\usepackage{multicol}
\usepackage{pslatex}
\usepackage{apalike}
\usepackage{array}
\usepackage{booktabs}
\usepackage{amsmath}
\usepackage{algorithm}
\usepackage{tabto}
\usepackage{bm}
\usepackage{tikz}
\usepackage{textcomp}
\usepackage[noend]{algpseudocode}
\usepackage{SCITEPRESS}     

\makeatletter
\def\BState{\State\hskip-\ALG@thistlm}
\makeatother
\usetikzlibrary{arrows,positioning,shapes.geometric}
\begin{document}

\subfigtopskip=0pt
\subfigcapskip=0pt
\subfigbottomskip=0pt

\title{Image Segmentation of Multi-Shaped Overlapping Objects}
%
\author{\authorname{Kumar Abhinav\sup{1\dagger}, Jaideep Singh Chauhan\sup{1\dagger} and Debasis Sarkar\sup{1}}
\affiliation{\sup{1}Department of Chemical Engineering, Indian Institute of Technology Kharagpur,Kharagpur, India}
\email{kumar.abhinav@iitkgp.ac.in, jaideepiit2@gmail.com, dsarkar@che.iitkgp.ernet.in }
}


\abstract{In this work, we propose a new segmentation algorithm for images containing convex objects present in multiple shapes with a high degree of overlap. The proposed algorithm is carried out in two steps, first we identify the visible contours, segment them using concave points and finally group the segments belonging to the same object. The next step is to assign a shape identity to these grouped contour segments. For images containing objects in multiple shapes we begin first by identifying shape classes of the contours followed by assigning a shape entity to these classes. We provide a comprehensive experimentation of our algorithm on two crystal image datasets. One dataset comprises of images containing objects in multiple shapes overlapping each other and the other dataset contains standard images with objects present in a single shape. We test our algorithm against two baselines, with our proposed algorithm outperforming both the baselines.}

\onecolumn \maketitle \normalsize \vfill

\section{\uppercase{Introduction}}
\label{sec:introduction}
\begin{figure}[!h]
\centering
\subfigure[Dataset-I Crystal.]{\epsfig{file = 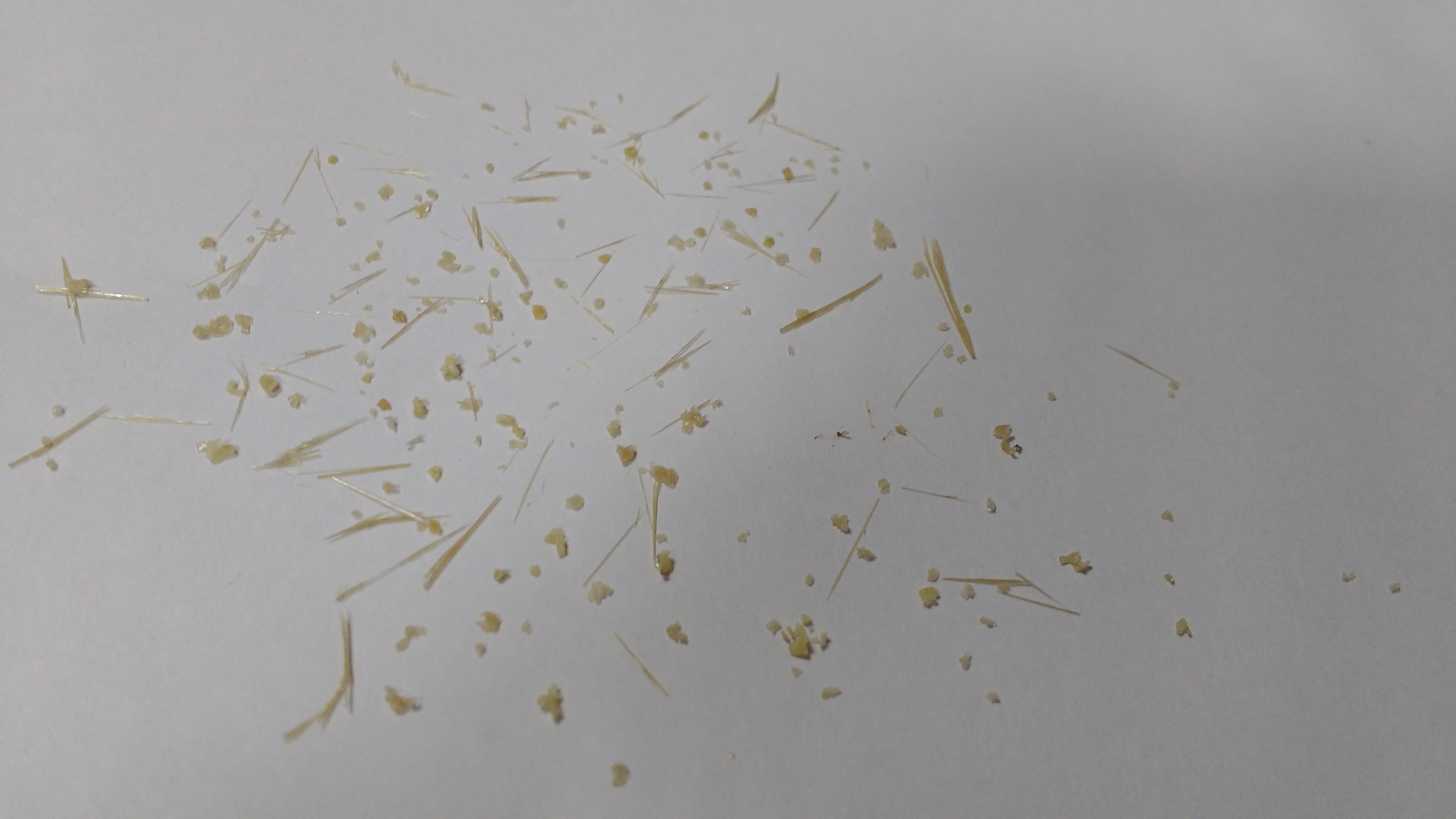, height = 3.5cm, width =3.5cm}}
\subfigure[Dataset-II Crystal.]{\epsfig{file = 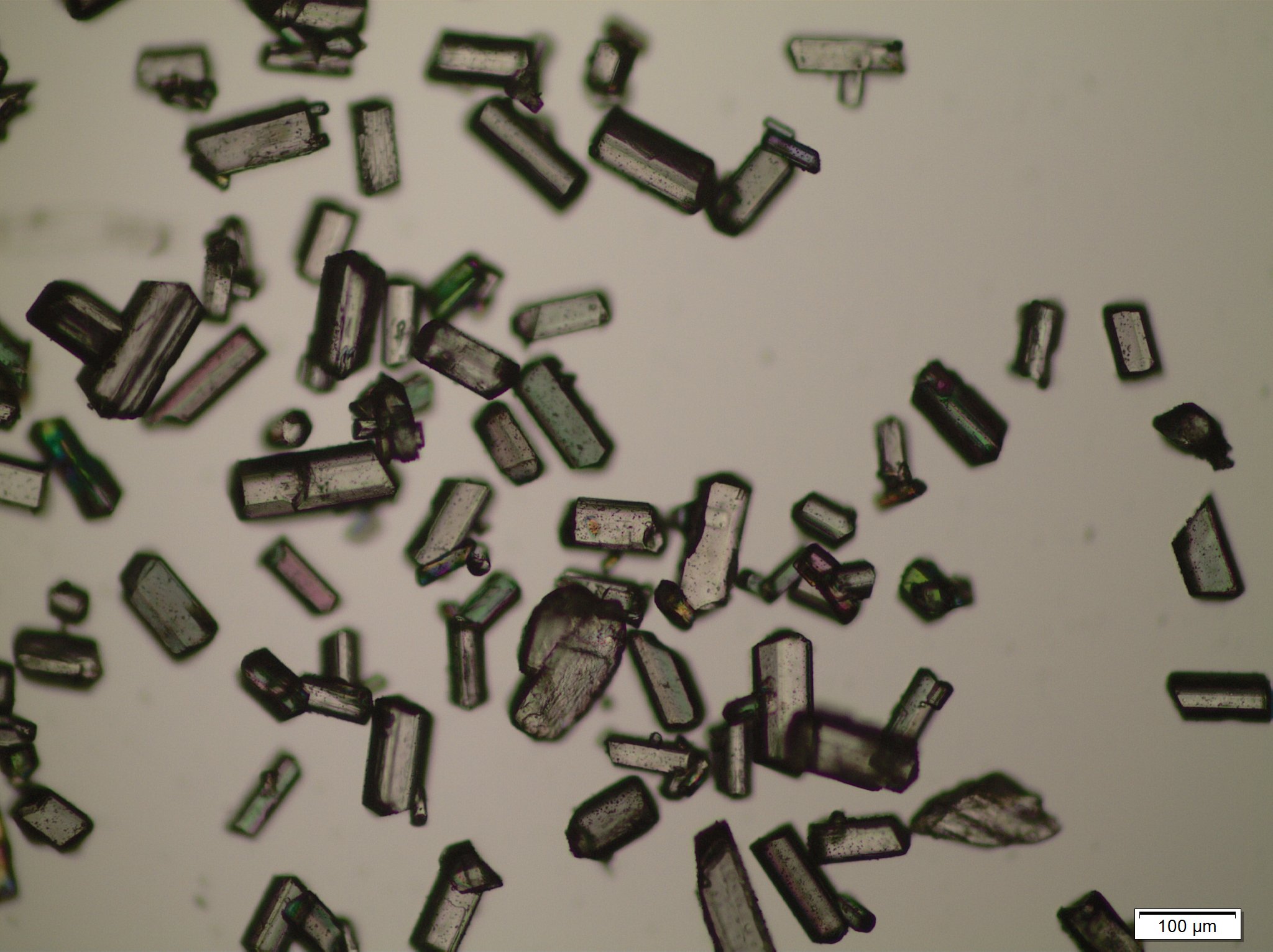, height = 3.5cm, width = 3.5cm}}
\caption{Overlapping Crystals Image.}\label{Fig1}
\end{figure}

\noindent A number of vision applications like Medical Imaging, Recognition Tasks, Object Detection, etc.\ require segmentation of objects in images. Overlapping objects in such images pose a major challenge in this task. Identifying hidden boundaries, grouping contours belonging to same objects and efficiently estimating the dimensions of partially visible objects are some of the complexities that overlap introduces. Further regions of high concentration of objects with a high degree of overlap introduces computational challenges for segmenting such images. In our work we introduce a method for tackling these issues for heterogeneous and homogeneous images. Heterogeneous images are defined as images containing multi-shaped overlapping objects, eg.\ a rod over a sphere. Homogeneous images are defined as images with similarly shaped overlapping objects. Segmentation is particularly challenging for a heterogeneous image because assigning a single shape entity to all the objects is bound to create inaccuracies. We test our work on two Datasets, Dataset-I (Fig.\ref{Fig1}.a) is a representative of the heterogeneous group of images, whereas Dataset-II (Fig.\ref{Fig1}.b) is a standard dataset containing homogeneous images.

To tackle the discussed challenges we break down the problem statement into the following three

\noindent\rule{4cm}{0.4pt} \linebreak
\tab $\dagger$ denotes equal contribution

sub-problems : (1) \textit{Contour extraction}, this corresponds to identifying the visible and hidden contours (2) \textit{Contour mapping}, this is for grouping the identified contours which belong to the same object and mapping them to it (3) \textit{Shape identification} for fitting a shape entity to the grouped contour segments, this step implicitly estimates the hidden parts. 

To address these issues we propose a two step algorithm. First, we identify the contours and break them into segments. This cleaving happens at all points where contours of two different objects intersect. These cleave points are identified using a procedure called concave point extraction described in Sec.3.2. Next we stitch the broken segments together which belong to the same objects. This is done by fitting an ellipse over all the broken segments, and stitching them together depending on their proximity and orientation. This concludes the first step and solves the first two sub-problems. The next step of the algorithm is to assign a shape entity to the mapped contours. This can be easily done for homogeneous images, a single shape entity like an ellipse fits all. It becomes tricky for heterogeneous image. The intuition of our work is that we need to treat each contour based on its shape class. So, first we define shape classes for the dataset. Followed by defining a specific shape entity for each class which can be applied to a contour that falls in it. The images in Dataset-I contains two kinds of objects, needle(or rod) shaped and spherical shaped, overlapping each other. We use these two shapes as the defined classes for assigning a shape entity. We fit an ellipse over the mapped segments and use its aspect ratio to determine the class to which the segment belongs to. Once classified each segment is treated according to the defined heuristic for the class. This step solves the third sub-problem and is followed by a post-processing step to reject the wrongly segmented objects.  

We summarize our major contributions as follows:
\begin{itemize}
\item We propose a new algorithm to segment heterogeneous images with regions of high object concentration and a high degree of overlap.
\item We test our work on both heterogeneous and homogeneous datasets, beating the baselines for both, demonstrating the general applicability of our work.
\end{itemize}

\section{\uppercase{Related Work}}
\label{sec:literature review}
\noindent Image segmentation has been in the research domain for a while, but the nature of problem is such that no approach generalizes well for a wide range of image datasets. Certain approaches extract edge information using the sharp changes in the pixel intensities in images \cite{pal1993review,rastgarpour2011application}. Object boundaries tend to introduce peaks in gradients of pixel intensity, thus these methods are effective in identifying the edges. Multiscale analysis \cite{gudla2008high} is one of the edge based methods which is robust enough to handle weak edges. Others use heuristic based grouping of pixels for identifying regions with similar properties in images. Region Growing \cite{adams1994seeded} and Region Splitting-Merging \cite{haralick1985image} are two of the common approaches applied in these set of methods. However, these approaches don't segment images with overlapping objects. Such images lack crucial edge and pixel information for the hidden parts which is required for the effectiveness of these approaches.  

Certain approaches use  dynamic programming for the task \cite{baggett2005whole}, they determine the most optimal boundaries for the hidden objects by defining a path of the highest average intensity along boundaries. Graph-cut methods segment objects by creating a graph out of pixels treated as nodes and difference between their intensities as a weight for edges, followed by finding the minimum cut for the graph \cite{shi2000normalized,felzenszwalb2004efficient}. Both of these approaches require prominent gradients at boundaries to be effective in segmenting objects. 

One of the popular approaches for segmentation of overlapping objects is the watershed algorithm \cite{vincent1991watersheds} which in its classical form often results in over-segmentation. A potential solution to over-segmentation is region merging, which does solve the problem but the method efficiency varies depending upon the object size and object distribution concentration in the image. Marker-controlled watershed \cite{malpica1997applying,yang2006nuclei} is also used for the purpose but the efficiency of the method depends highly on the accurate identification of the markers.

\cite{zhang2006segmenting} formulates the problem as a combination of level set functions and uses similarity transforms to predict the hidden parts of partially visible objects. The drawback of this method is that its performance is dependent on initialization and is computationally expensive. \cite{alvarez2010morphological} proposes a morphological extension to the level based methods, it introduces morphological operations in traditional snake algorithms. This method outperforms earlier active contour methods as the curve evolution is simpler and faster. It however fails to converge when large number of closely spaced overlapped objects are present in the image. In \cite{zhang2012method}, the problem of segmenting overlapping elliptical bubbles is solved by segmenting the extracted contours into curves. These curves are further segmented into groups belonging to the same bubble followed by ellipse fitting. This approach works well for images having similar shaped objects, but fail for images containing multi-shaped overlapping objects.         
    


Our proposed segmentation algorithm is inspired from concave point based overlapping object segmentation method \cite{zafari2015segmentation}. In \cite{zafari2015segmentation}, the idea of grouping the contours belonging to the same object present in an overlap cluster is carried out by fitting ellipses over the segmented contours. The grouped segments are further fitted with ellipses to separate overlapping objects. The efficiency of this approach declines when applied to images with multi-shaped objects.

Our approach is built for images containing objects with high degree of overlap and we attempt to solve the following issues present in earlier discussed approaches: (1) We attempt to segment images with regions of high object concentration, which becomes a problem for approaches which are computationally expensive, (2) Our approach does not rely on strong gradients at boundaries to be effective and (3) Most approaches don't take into account images with overlapping multi-shaped objects, we use adaptive shape fitting to address this problem.

\section{\uppercase{Proposed Methodology}}
\label{sec:methods}

\begin{figure}[!h]
\centering
{\epsfig{file = 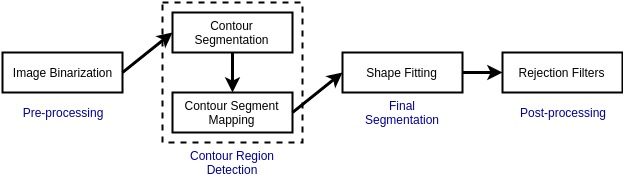, width = 7.5cm}}
\caption{Overview of the proposed approach.}\label{Fig2}
\end{figure}

\noindent We propose a segmentation algorithm consisting of the following two steps: Contour Region Detection (CRD) and Shape Fitting. Fig.\ref{Fig2} describes the methodology. 

\begin{figure}[!h]
\centering
\subfigure[Dataset-I image.]{\epsfig{file = 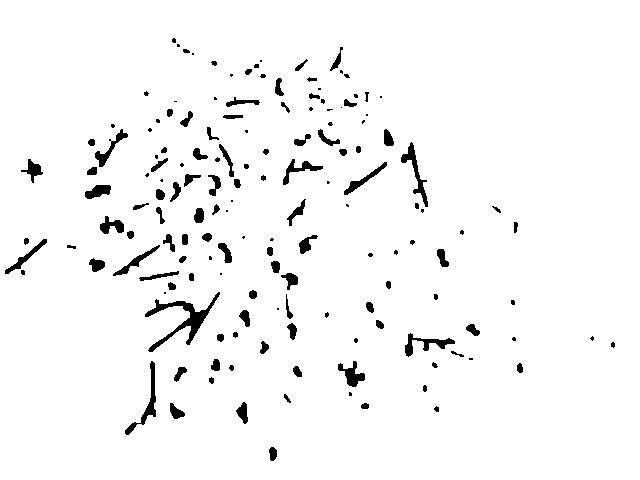, height = 3.5cm, width =3.5cm}}
\subfigure[Dataset-II image.]{\epsfig{file = 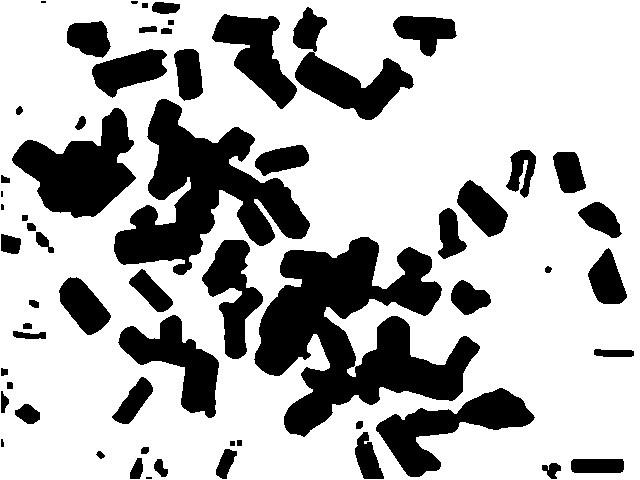, height = 3.5cm, width = 3.5cm}}
\caption{Binarized Image for both the Datasets.}\label{Fig3}
\end{figure}

We take in RGB images as input and convert them to grayscale before further processing. After conversion, median blur is applied to remove noise from the image, followed by Image binarization. This is carried out using Otsu's method \cite{otsu1979threshold} from which we get a binary image $B_O$. Finally, morphological operations are used to filter noisy regions and fill any small holes in the binary cell regions of $B_O$. Fig.\ref{Fig3}.a and \ref{Fig3}.b shows the binarized image of Fig.\ref{Fig1}.a and Fig.\ref{Fig1}.b respectively.

CRD step is then carried out on the binary image, $B_O$ to extract the contour information. It identifies the contours corresponding to both visible and hidden boundaries and return contour segments grouped together on basis of them belonging to the same object. Shape Fitting is the final segmentation step where a shape entity like an ellipse or a rod is assigned to the grouped contour segments. Rejection Filters are applied in the post-processing step for identifying and eliminating wrongly segmented objects.

\subsection{Contour Region Detection}
\noindent CRD is the first step of our proposed methodology. In this step, segmentation of overlapping objects is carried out using the boundaries of the visible objects. We begin first by extracting the contour points corresponding to each of the object's contour present in the image. For this we use the border following algorithm described in \cite{suzuki1985topological}. Next we describe some notations that we'll be using from hereon, let the detected object's contour $C$ is described using a set $C_{contour}$ with $N$ contour points such that, $C_{contour}$ = \{$p_{i}$($x_{i}$,$y_{i}$); $i$=1,2,...,$N$\}, where $N$ is the number of contour points and $p_{i}$ is the $i^{th}$ contour point with coordinates $(x_{i}, y_{i})$. $p_{i+1}$ and $p_{i-1}$ are the neighbouring contour points of $p_{i}$. The contour region detection next involves two distinctive tasks: contour segmentation and contour segment mapping. 

\subsubsection{Contour Segmentation}
\begin{figure}[!h]
\centering
{\epsfig{file = 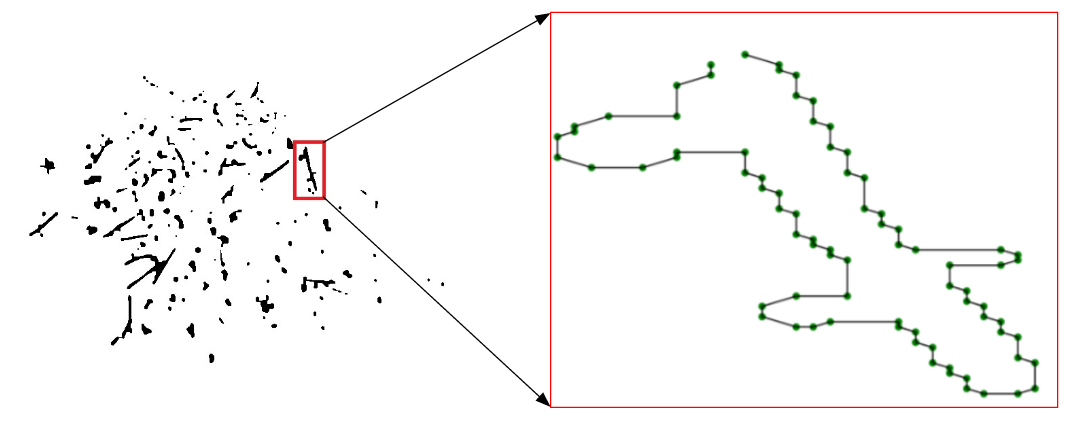, , width =7cm}}
\caption{Plotting a contour from Dataset-I to show its Contour points, which are represented using Green points.}\label{Fig4}
\end{figure}

\noindent The contour points extracted previously represents the contour of a single object or multiple overlapped objects being detected as a single entity. Fig.\ref{Fig4} displays the detected contour points on a sample contour of the binarized crystal image shown in Fig.\ref{Fig3}.a. The overlapping of objects forms an irregular closed shape. The intersections of the object boundaries in a overlapped object contour are represented using break points or concave points. Concave points are thus used to segment the contour of overlapping objects into separate contour segments. 

The concave points are determined by a two step procedure: (1) Determining the corner points using RDP algorithm and (2) Identifying concave points from corner points using convexity check. 

\begin{figure}[!h]
\centering
\subfigure[Corner Points Detection (Red points).]{\epsfig{file = 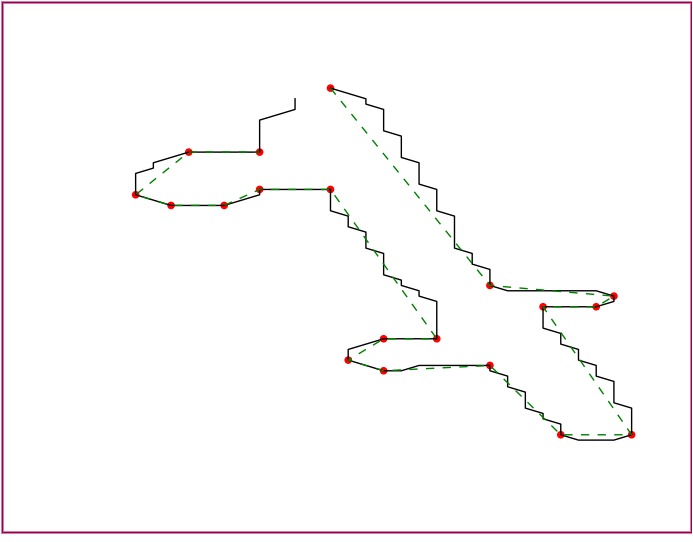, height = 3.5cm, width =3.5cm}}
\subfigure[Concave Points Detection (Red points).]{\epsfig{file = 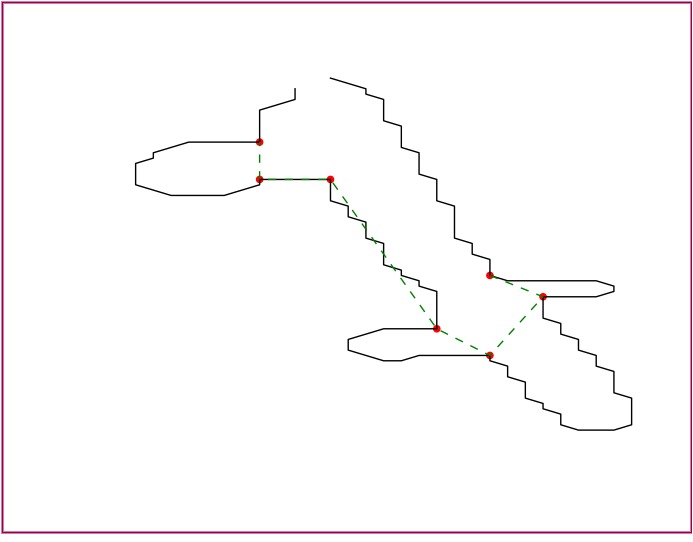, height = 3.5cm, width = 3.5cm}}
\caption{Contour segmentation points of a sample contour in Dataset-I.}\label{Fig5}
\end{figure}

The corner points are the points of local minima or maxima of the curve drawn by plotting the predetermined contour points. Thus, for each of the object's contour governed by its contour points, the corner points are computed using Ramer-Douglas-Peucker (RDP) algorithm \cite{douglas1973algorithms}. RDP is a path simplification algorithm which reduces the number of contour tracing points. The RDP algorithm takes in an ordered sequence of contour points represented using set $C_{contour}$ as input. The algorithm returns a set of corner points extracted from the set $C_{contour}$. Let this set of corner points be represented using a set $C_{corner}$, $C_{corner}$ = \{$p_{cor,i}$($x_{i}$,$y_{i}$) $\mid$ $p_{cor,i}$ $\in$ $C_{contour}$; $i$ = 1,2,...,$M$\}, where $p_{cor,i}$ is the $i^{th}$ corner point and $M$ be number of corner points extracted such that $M \leq N$. Fig.\ref{Fig5}.a shows a set of corner points being extracted from contour points of a sample contour as displayed in Fig.\ref{Fig4}. 

Convexity check is then carried out on the set of corner points contained in $C_{corner}$ to obtain a set of concave points. The algorithm to compute concave points is described in Algorithm \ref{algo1}.

In Algorithm \ref{algo1}, detected corner points from $C_{corner}$ are the inputs. Initially vectors are created by joining two adjacent corner points in clockwise direction along direction of second corner point. Thus, each corner point will be contained in two vectors, one originating from the concerned point and the other terminating at it. For every corner point, the sign of the cross product of the two vectors it is a part of, is computed and stored as the orientation of that corner point. Net orientation of the contour is then computed by taking sign of sum of all the orientations, which basically gives effective sign over all orientations. If the net orientation is positive it means that the overall contour traversal along the corner points is in anti-clockwise direction else in clockwise direction. Wherever the corner point's orientation is different from the net orientation i.e.\ there is a break in the traversal direction, that corner point is marked as a concave point. 

\begin{algorithm}
\fontsize{10pt}{10pt}\selectfont
\caption{Concave Point Extraction}\label{algo1}
\begin{algorithmic}[1]
\State net\_orient = 0 \Comment{Overall orientation of contour}
\State orientation = []\Comment{Orientation at each corner point} 
\State FOR  $i$ = 1 to $M$ : \Comment{$M$ is \# Corner Points}
\State \tab GET $p_{cor,i}$ , $p_{cor,i-1}$ , $p_{cor,i+1}$ $\in$ $C_{corner}$ 
\State \tab IF $i$ \textit{equals} 1 :
\State \tab \qquad $p_{cor,i-1}$ = $p_{cor,M}$
\State \tab ENDIF
\State \tab IF $i$ \textit{equals} $M$ :
\State \tab \qquad $p_{cor,i+1}$ = $p_{cor,1}$
\State \tab ENDIF
\State \tab \Comment{Vector initialization from given two points}
\State \tab $\bm{V_{i}}$ = ($p_{cor,i-1}$ , $p_{cor,i}$) 
\State \tab $\bm{V_{i+1}}$ = ($p_{cor,i}$ , $p_{cor,i+1}$)
\State \tab orientation[$i$] $\gets$ SIGN($\bm{V_{i}}$ $\times$ $\bm{V_{i+1}}$)
\State \tab net\_orient $\gets$ net\_orient + orientation[$i$]
\State ENDFOR
\State net\_orient $\gets$ SIGN(net\_orient) 
\State \Comment{If net\_orient is "+ve" means anti-clockwise if "-ve" means clockwise}
\State FOR  $i$ = 1 to $M$ :
\State \tab IF orientation[$i$] \textit{not equals} net\_orient :
\State \tab \Comment{Break in traversal direction}
\State \tab \qquad INSERT $p_{cor,i}$ \textit{in} $C_{concave}$
\State \tab ENDIF
\State ENDFOR
\end{algorithmic}
\end{algorithm} 

Let the set of concave points computed from Algorithm \ref{algo1} be represented using a set $C_{concave}$, $C_{concave}$ = \{$p_{concave,i}$($x_{i}$,$y_{i}$) $\mid$ $p_{concave,i}$ $\in$ $C_{corner}$; $i$ = 1,2,...,$P$\}, where $p_{concave,i}$ is the $i^{th}$ concave point and $P$ be number of concave points computed such that $P <= M <= N$. Fig.\ref{Fig5}.b displays the concave points being extracted from the corner points of a sample contour displayed in Fig.\ref{Fig5}.a. 

\subsubsection{Contour Segment Mapping}

\noindent The detected concave points in set $C_{concave}$ are used to split the earlier detected object's contour $C$ into contour segments $L_{i}$'s. Let $L_{i}$ = \{$p_{i_1}$,$p_{i_2}$,...,$p_{i_s}$\} represent a contour segment of the contour $C$ where $s$ is the number of contour points on $L_{i}$. The set $L_i$ contains the set of contour points from $C_{contour}$ present between two concave points, $p_{i_1}$ and $p_{i_s}$. The start and end points of the corresponding segment, $L_i$ are represented by $p_{i_1}$ and $p_{i_s}$, such that $p_{i_1}$ ,$p_{i_s}$ $\in$ $C_{concave}$. Since total number of concave points are $P$, we have 
\begin{equation}\label{eq1}
    C = L_{1} + L_{2} + ... + L_{P}
\end{equation}
From now on, the detected contours $C$ are referred to as contour clusters. A contour cluster is a general term for detected contours which may contain multiple overlapped objects or just a single object. Our proposed segmentation algorithm segments the overlapped objects present in a contour cluster, if overlapping exists otherwise it just segments the single object present. The sample contour present in Fig.\ref{Fig4} can now be referred as a contour cluster.

Overlapping of objects and their shape irregularities in a contour cluster lead to a presence of multiple contour segment of a single object. Contour segment mapping is thus needed to map together all the contour segments belonging to a same object in a contour cluster.

The segment mapping algorithm iterates over each pair of the contour segment($L_{i}$'s), checking if they could be mapped (grouped) together. The algorithm follows orientation based mapping, where an ellipse is first fitted on each of the contour segment using classical least square fitting. The fitted ellipse on a contour segment is then mapped with other ellipses based on proximity and orientation. The algorithm for segment mapping is discussed in Algorithm \ref{algo2}. 

\begin{algorithm}
\fontsize{10pt}{10pt}\selectfont
\caption{Contour Segment Mapping}\label{algo2}
\begin{algorithmic}[1]
\State \Comment{$L_i^{’s}$ : $i^{th}$ contour segment of contour cluster $C$}
\State FOR $i$ = 1 to $P$ : \Comment{$P$ is \# Concave Points}
\State \tab \textit{fit ellipse} $E_{i}$ \textit{to segment} $L_{i}$
\State \tab $major\_axis_i$,$minor\_axis_i$,$orient\_angle_i$ $\gets$ $E_i$
\State \tab $aspect\_ratio_i$ $\gets$ $\frac{major\_axis_i}{minor\_axis_i}$
\State \tab IF $aspect\_ratio_i$ $<$ 2 :
\State \tab \qquad \Comment{Fitted ellipse close to a circle}
\State \tab \Comment{Mark as unique segment, isn't mapped}
\State \tab \qquad MAP $L_i$ \textit{alone to} Set $G$
\State \tab \qquad CONTINUE
\State \tab ENDIF
\State \tab FOR $j$ = $i+1$ to $P$ :
\State \tab \qquad \textit{fit ellipse} $E_{j}$ \textit{to segment} $L_{j}$
\State \tab \qquad $major\_axis_j$, $minor\_axis_j$  $\gets$ $E_j$
\State \tab \qquad $orient\_angle_j$ $\gets$ $E_j$
\State \tab \qquad $aspect\_ratio_j$ $\gets$ $\frac{major\_axis_j}{minor\_axis_j}$
\State \tab \qquad IF $|$ $orient\_angle_i$ - $orient\_angle_j$ $|$ $<$ $\epsilon$:
\State \tab \qquad \qquad MAP \textit{together} $L_i$ \textit{and} $L_j$ \textit{to} Set $G$ 
\State \tab \qquad ENDIF
\State \tab ENDFOR
\State ENDFOR
\end{algorithmic}
\end{algorithm} 

In Algorithm \ref{algo2}, the contour segment with fitted ellipse close to a circle (aspect ratio $\leq$ 2) isn't mapped with any other contour segment. It's assumed that only elongated ellipses, with larger aspect ratio ($>$ 2) have more than one contour segment present in the contour cluster $C$. 


For the contour cluster $C$ containing $P$ contour segments, let these segments are mapped to form $K$ groups. This new unique contour set containing groups of mapped together contour segments is represented using a set $G$ with $K$ elements such that : 
\begin{equation}\label{eq2}
    G = \{ G_{1},G_{2},...,G_{K} \}
\end{equation}
\begin{equation}\label{eq3}
    \bigcup_{i=1}^{K} G_{i} = \{ L_{1},L_{2},...,L_{P} \}
\end{equation}
\begin{equation}\label{eq4}
    \bigcap_{i=1}^{K} G_{i} = \Phi
\end{equation}
An element in this unique set of grouped contour segments $G$, represents the entire visible contour of an unique object present in the overlapping object contour cluster. 

\begin{figure}[!h]
\centering
\subfigure[Before Contour Segment Mapping.]{\epsfig{file = 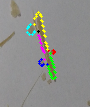, height = 3.5cm, width =3.5cm}}
\subfigure[After Contour Segment Mapping.]{\epsfig{file = 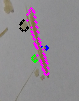, height = 3.5cm, width = 3.5cm}}
\caption{Contour Segment Mapping result on a sample contour cluster from Dataset-I(colors are used for illustration to visualize contour segment mapping, images are magnified thus continuous color segments looks broken).}\label{Fig6}
\end{figure}

Fig.\ref{Fig6} displays the result of the Contour Segment Mapping using color visualization. The sample contour cluster is the one shown in Fig.\ref{Fig4}. Fig.\ref{Fig6}.a shows the result before segment mapping, where each colored segment represents a contour segment $L_{i}$ of the above mentioned sample contour cluster, $C$. These segments are split using concave points as displayed in Fig.\ref{Fig5}.b. Fig.\ref{Fig6}.b shows the result after segment mapping. It can be seen that earlier present contour segments denoted by yellow, green and pink colors respectively now gets mapped together and is represented using pink color. Each of the unique colored segments in Fig.\ref{Fig6}.b represents an element of the set $G$. 

\subsection{Shape Fitting}

\begin{figure}[!h]
\centering
\subfigure{\epsfig{file = 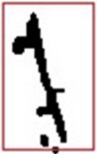, height = 3.5cm, width = 2cm}}
\subfigure{\epsfig{file = 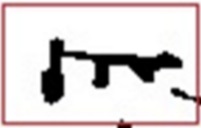, height = 3.5cm, width = 3cm}}
\subfigure{\epsfig{file = 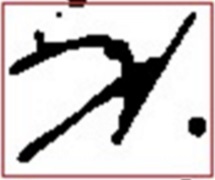, height = 3.5cm, width = 2cm}}
\subfigure{\epsfig{file = 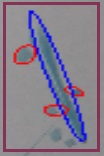, height = 3.5cm, width = 2cm}}
\subfigure{\epsfig{file = 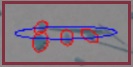, height = 3.5cm, width = 3cm}}
\subfigure{\epsfig{file = 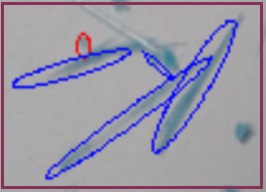, height = 3.5cm, width = 2cm}}
\caption{Overlapped object segmentation in Dataset-I. First row contains three different detected contour clusters, second row shows segmentation of corresponding clusters.}\label{Fig7}
\end{figure}

\noindent The final step of the algorithm is to assign a shape entity to the object contours present in a contour cluster. Here we present a heuristic based approach for assigning this entity. We begin with modeling the contours with ellipse fitting. The classic least square fitting algorithm is used for assigning the ellipses. An ellipse is fitted on each element of the set $G$ governed by Eq.\ref{eq2}-\ref{eq4}(representative of all the contours belonging to the same object). Fig.\ref{Fig7} displays result of the segmentation on some detected contour clusters of the binarized crystal image(Fig.\ref{Fig3}.a).

\begin{figure}[!h]
\centering
\subfigure[Normal ellipse fitting.]{\epsfig{file = 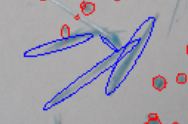, height = 3.5cm, width =3.6cm}}
\subfigure[Bounding box algorithm.]{\epsfig{file = 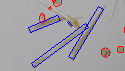, height = 3.5cm, width = 3.8cm}}
\caption{Rod shape crystal estimation using bounding box.}\label{Fig8}
\end{figure}

For Dataset-II, ellipse fitting suffices the segmentation process, as it contains rectangular shaped objects whose dimensions are well represented by the major and minor axes of the fitted ellipse. However, the heuristics are required for Dataset-I, where the contours are needed to be classified according to their shape representation. The aspect ratio of the fitted ellipse determines its closeness to a circle (aspect ratio $<$ 2) or a rod. The threshold value for the aspect ratio is set to be 2 by analyzing the data manually. We cannot represent the rod shaped crystals with ellipse due to a problem of overestimation of length. This is due to the fact that the rod shaped crystals have very small width and the fitted ellipse couldn't take account of this. As a solution to this, we propose a new bounding box algorithm as a shape estimator for rod shaped crystals. It takes into account the small width of the crystal. The results after applying this algorithm are shown in Fig.\ref{Fig8}. The procedure for the bounding box algorithm is discussed below:
\begin{enumerate}
  \item The orientation angle, center coordinates and the major and minor axis of the fitted ellipse on the rod shaped crystal's contour are computed.
  \item A pair of parallel line is used with their slope equal to the ellipse's orientation angle and their initial position coinciding with the major axis.
  \item The parallel lines are separated equidistantly from the major axis along the direction of minor axis in opposite directions.
  \item The separation distance is increased until a far end of the crystal contour is achieved at the both side.
  \item The distance between parallel lines gives the width of the rod shaped crystal.
  \item The same algorithm is extended to calculate rod's length using a new parallel line pair with initial slope along normal to orientation angle and initial position coinciding with minor axis.
  \item The bounding box generated by the intersection of these two pairs of parallel lines at their final positions gives the desired shape estimator for the segmentation of rod shaped crystals. 
\end{enumerate}

The calculated dimensions are converted from pixels to micrometres using scaling factor of the microscope and the camera. 

\subsection{Rejection Filters}
\noindent This is the post-processing step of our proposed methodology. It reduces the false positive segmentation which may be introduced due to incorrect ellipse fitting in the shape fitting step. The two filters being used are discussed accordingly. 

\begin{figure}[!h]
\centering
{\epsfig{file = 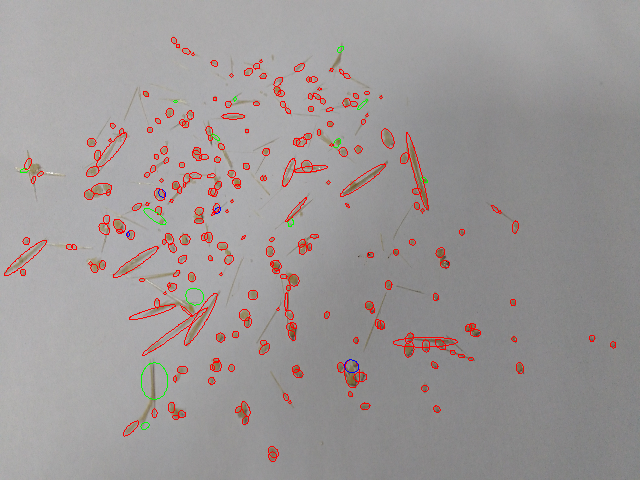, , width =5cm}}
\caption{Effect of Rejection Filter on Dataset-I. Red ellipses are final fitted ellipse after rejection filters, Green ellipses are rejected due to \textit{masking rejection} filter, Blue ellipses are rejected due to \textit{area overlap} filter.}\label{Fig9}
\end{figure}

The first filter is a \textit{masking rejection} filter, which checks whether the fitted ellipse covers at least 75\% black pixels within its boundaries in the binary image. It makes sure that the fitted ellipse is not empty.
Finally, an \textit{area overlap} filter is used, which on encountering a high degree of overlap between neighbouring ellipses rejects the neighbouring smaller ellipse. A high degree of overlap is defined by the larger ellipse encompassing more than 50\% of the area of the smaller ellipse. The output of applying different rejection filters is displayed in Fig.\ref{Fig9}. 

\section{\uppercase{Experiments}}
\label{sec:experiments}
\subsection{Dataset}
\noindent We choose two datasets to test our proposed algorithm. The first dataset (Dataset-I, Fig.\ref{Fig1}.a), comprise images taken from a hand-held camera of crystals present in rod and circular shapes. It consists of 2 subsets of 10 images, each image has an average of 200 crystals. Each of these two subsets contains different percentages of rod and circular shaped crystals. The second dataset (Dataset-II, Fig.\ref{Fig1}.b), consists of microscopic images of similarly shaped rectangular crystals. It consists of 15 images, each image has an average of 100 crystals. The images in both datasets are of dimension 640 x 480 pixels. The ground truth is generated by manually marking crystal's center in all the images contained in both datasets. Dataset-II represents a generic homogeneous dataset that most existing algorithms work with, such datasets contain similarly shaped crystals overlapping each other. On the other hand Dataset-I is a heterogeneous dataset, containing multi-shaped crystals overlapping each other. We demonstrate the novelty of our proposed algorithm on this dataset. 

\subsection{Performance Measures}
\noindent To evaluate our algorithm performance, first we compare our results with two baselines followed by verification of results with the experimental crystal mixture data. For evaluation we use precision and recall as follows:

\begin{equation}\label{eq6}
   Precision = \frac{TP}{TP + FP}
\end{equation}

\begin{equation}\label{eq6}
   Recall = \frac{TP}{TP + FN}
\end{equation}

where TP is True Positive and it is defined the number of correctly segmented crystals, False Negative (FN) is the number of unsegmented crystals and False Positive (FP) is the number of incorrectly segmented crystals. The values of TP, FN and FP are calculated manually and are taken average over each of the datasets.

As a sanity check we use Jaccard Similarity Coefficient (JSC) to check the accuracy of our segmentation \cite{choi2010survey}. A binary map of the segmented object $O_s$ and the ground truth particle $O_g$ is created. Distance between the ground truth particle's center and the fitted ellipse's (or rod's) center for a segmented particle is used as the parameter for binary map generation. The JSC is computed as follows:

\begin{equation}\label{eq6}
   JSC = \frac{n(O_s \cap O_g)}{n(O_s \cup O_g)}
\end{equation}

The distance threshold value (JSC threshold, $\rho$) for the binary map is set to 8 pixels. It means that if the distance between the ground truth crystal's center and fitted ellipse's center on segmented crystal is less than 8 pixels then they are mapped together in the binary map i.e.\ added to the $O_s \cap O_g$ set. 

\begin{figure}[!h]
\centering
{\epsfig{file = 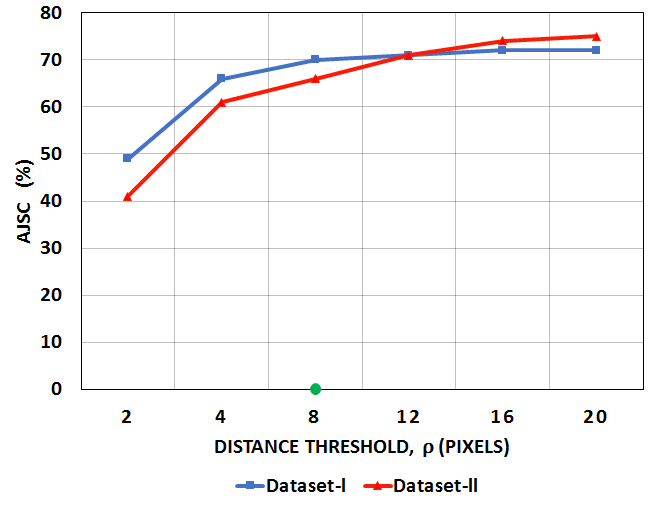, , width =5cm}}
\caption{AJSC performance of the proposed segmentation method with different values of distance threshold, $\rho$.}\label{graph}
\end{figure}

The average JSC (AJSC) value over each dataset is thus used as a third measure to evaluate the segmentation performance. Fig.\ref{graph} presents the variation of AJSC value of our proposed algorithm with different values of threshold $\rho$ for both the datasets. Note, here a higher value of $\rho$ represents a higher leniency in measuring the accuracy for segmentation. Our algorithm achieves very similar AJSC values for both the datasets for low $\rho$ values.

\subsection{Results}
\begin{figure}[!h]
\centering
\subfigure[Blue boxes represent rod crystals, Red ellipses represent circular crystals.]{\epsfig{file = 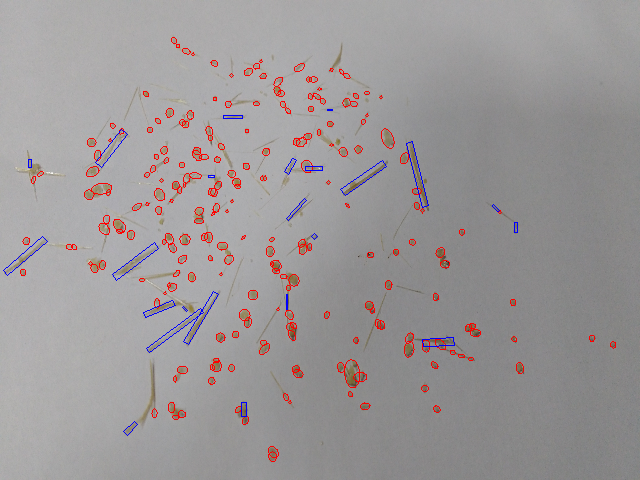, height = 3.5cm, width =3.5cm}}
\subfigure[Red ellipses represent similarly shaped crystals.]{\epsfig{file = 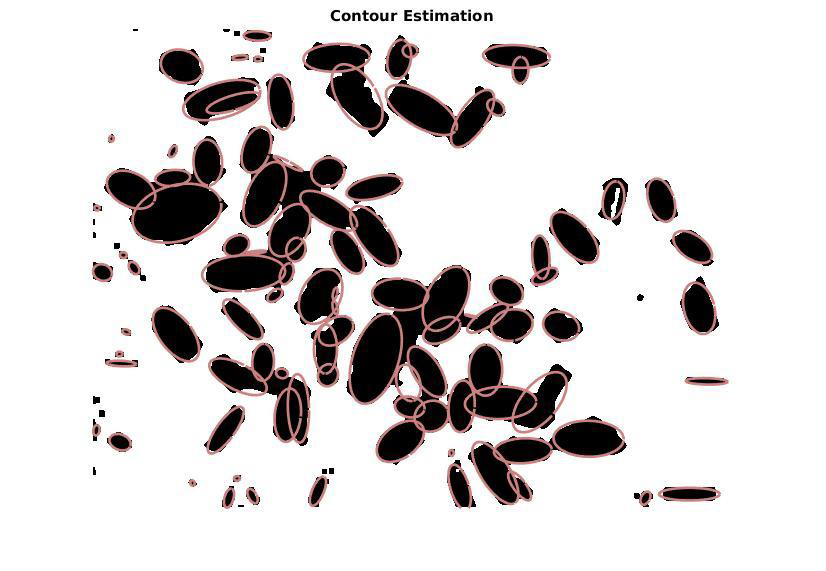, height = 3.5cm, width = 3.5cm}}
\caption{Final segmentation result of our proposed algorithm on sample image from Dataset-I(a) and Dataset-II(b).}\label{Fig10}
\end{figure}

\noindent Fig.\ref{Fig10}.a and Fig.\ref{Fig10}.b show results of our proposed segmentation algorithm implemented on the two different set of crystal image samples, Fig.\ref{Fig1}.a and Fig.\ref{Fig3}.b respectively.

The performance of the proposed segmentation method is compared with two existing segmentation algorithms, Segmentation of Partially Overlapping Nanoparticles (SPON) \cite{zafari2015segmentation} and Segmentation of Overlapping Elliptical Objects (SOEO) \cite{zafari2015segmentation-b}. The SPON and SOEO methods are particularly chosen as it is previously applied for segmentation of overlapping convex objects being represented by an elliptical shape. The implementation made by the corresponding authors is used for SPON \cite{zafari2015segmentation} and SOEO \cite{zafari2015segmentation-b}.

\begin{figure}[!h]
\centering
\subfigure[Proposed]{\epsfig{file = 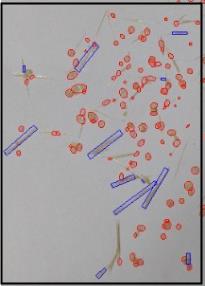, height = 3.5cm, width = 2.4cm}}
\subfigure[SPON]{\epsfig{file = 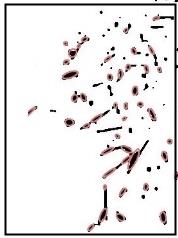, height = 3.5cm, width =2.4cm}}
\subfigure[SOEO]{\epsfig{file = 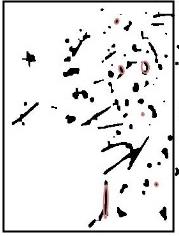, height = 3.5cm, width =2.4cm}}
\caption{Dataset-I segmentation comparison. Results of proposed algorithm is displayed on original crystal image whereas SPON and SOEO on binarized crystal image.}\label{Fig11}
\end{figure}

\begin{figure}[!h]
\centering
\subfigure[Proposed]{\epsfig{file = 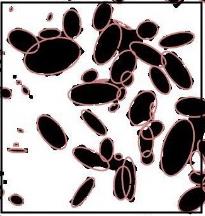, height = 3.5cm, width = 2.4cm}}
\subfigure[SPON]{\epsfig{file = 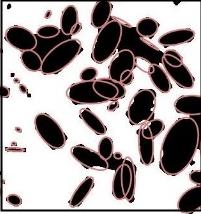, height = 3.5cm, width =2.4cm}}
\subfigure[SOEO]{\epsfig{file = 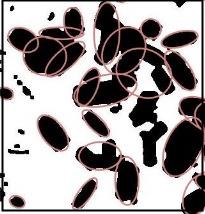, height = 3.5cm, width =2.4cm}}
\caption{Dataset-II segmentation comparison. All the results are displayed on the binarised crystal image.}\label{Fig12}
\end{figure}

Examples of visual comparison of the segmentation results for both the datasets are presented in the Fig.\ref{Fig11} and \ref{Fig12}. Distinctive segmentation efficiency of our proposed algorithm could be observed as compared to SPON and SOEO. Binarized image after pre-processing is used as an input to the SPON and SOEO algorithm since the pre-processing step is dependent on the experimental setup.  

The corresponding performance statistics of the competing methods applied to the two datasets are presented in Table\ref{table1} and Table\ref{table2}. Our algorithm outperforms both SPON and SOEO in terms of Recall and AJSC criterias for both the datasets whereas comparable results are obtained for Precision criteria. For Dataset-I the gain is substantial with our algorithm outperforming SPON by $36\%$ on Recall. SOEO gives a very poor score for Dataset-I ($2\%$ Recall) reaffirming the challenges of segmenting images with multi-shaped objects. We also achieve substantially higher AJSC values for both the datasets indicating improved accuracy of placing segments in the image for our algorithm. $\rho$ is taken to be 8 $pixels$ for computing the AJSC value.

\begin{table}[h]
\caption{Dataset-I performance comparison.}\label{table1} \centering
\begin{tabular}{|c|c|c|c|}
  \hline
  Methods & Recall \% &Precision \% & AJSC \%\\ 
  \hline
  Proposed & \textbf{87} & 85 & \textbf{70} \\
  \hline
   SPON & 51 & \textbf{89} & 28\\
   \hline
   SOEO & 02 & 75 & 04 \\
  \hline
\end{tabular}
\end{table}

\begin{table}[h]
\caption{Dataset-II performance comparison.}\label{table2} \centering
\begin{tabular}{|c|c|c|c|}
  \hline
  Methods & Recall \% & Precision \% & AJSC \%\\
  \hline
  Proposed & \textbf{82} & \textbf{87} & \textbf{66}\\
  \hline
   SPON & 81 & 82 & 31\\
  \hline
   SOEO & 39 & 76 & 23\\
   \hline
\end{tabular}
\end{table}

We provide a verification of our segmentation algorithm using composition analysis. Dataset-I contains images of the crystal p-Amino benzoic acid which is present in two shapes namely circular (prism) and the rod (needle). The circular shaped crystal is called the $\beta$-polymorph whereas the rod shaped crystal is called the $\alpha$-polymorph. Using known dimensions of the two shapes from Section 3.2, the volumes are computed by estimating rods as cylinders and circular crystals as spheres. Weight percentage of each of these polymorphs are then computed using known density, 1.369 g/ml and 1.379 g/ml for $\alpha$-polymorph and $\beta$-polymorph respectively.

Two sets of crystal samples are prepared by mixing different known weight percentage of each of the polymorph. Using our algorithm, for each of the set, average weight percentage of each shape is computed and checked against known weight percentage of the two samples. The corresponding performance statistics for the verification step is shown in Table \ref{table3}. The algorithm has an average error of 1.95\% over the actual weight percentages.

\begin{table}[h]
\caption{Verification from weight analysis, Actual : known sample percentage; Image: calculated weight percentage from proposed algorithm.}\label{table3} \centering
\begin{tabular}{*5c}
\toprule
Polymorphs  & \multicolumn{2}{c}{Sample A} & \multicolumn{2}{c}{Sample B}\\
\midrule
{}                  & Actual     & Image     & Actual      & Image\\
$\alpha$ wt.(\%)      &  0.0    & 3.5    & 20.0     & 19.6\\
$\beta$ wt.(\%)       &  100.0  & 96.5   & 80.0     & 80.4\\
\bottomrule
\end{tabular}
\end{table}

Finally, we do a complexity analysis. For $K$ detected contour points, $M$ corner points and $P$ concave points the algorithm has the following complexity :

\begin{equation}
\mathcal{O}(K*(M+P^2))
\end{equation}

The proposed method is implemented in Python, using a PC with a 2.60 GHz CPU and 6GB of RAM. The computational time of our proposed algorithm is around 0.5s per input image of size 640 x 480 pixels.




\section{\uppercase{Conclusions}}
\label{sec:conclusions}
\noindent A new method for segmenting convex objects with high degree of overlap is developed in this paper. The algorithm uses concave point for segmentation and heuristic based approach for adaptive shape fitting. The key novelty of our work is to segment multi-shaped convex objects with high degree of overlap present in high density in an image. The proposed algorithm is tested on two different datasets, the first dataset contains multi-shaped crystals and the other contains similar shaped crystals. The algorithm performance is compared against two competing algorithms, SPON and SOEO, with our proposed algorithm outperforming both.

\bibliographystyle{apalike}
{\small
\bibliography{main}}

\end{document}